# Channels' Confirmation and Predictions' Confirmation: from the Medical Test to the Raven Paradox

Chenguang Lu

survival99@gmail.com

**Abstract:** After long arguments between positivism and falsificationism, the verification of universal hypotheses was replaced with the confirmation of uncertain major premises. Unfortunately, Hemple discovered the Raven Paradox (RP). Then, Carnap used the logical probability increment as the confirmation measure. So far, many confirmation measures have been proposed. Measure $F$ among them proposed by Kemeny and Oppenheim possesses symmetries and asymmetries proposed by Elles and Fitelson, monotonicity proposed by Greco *et al.*, and normalizing property suggested by many researchers. Based on the semantic information theory, a measure $b^*$ similar to $F$ is derived from the medical test. Like the likelihood ratio, $b^*$ and $F$ can only indicate the quality of channels or the testing means instead of the quality of probability predictions. And, it is still not easy to use $b^*$, $F$, or another measure to clarify the RP. For this reason, measure $c^*$ similar to the correct rate is derived. The $c^*$ has the simple form: $(a-c)/\max(a, c)$; it supports the Nicod Criterion and undermines the Equivalence Condition, and hence, can be used to eliminate the RP. Some examples are provided to show why it is difficult to use one of popular confirmation measures to eliminate the RP. Measure $F$, $b^*$, and $c^*$ indicate that fewer counterexamples' existence is more essential than more positive examples' existence, and hence, are compatible with Popper's falsification thought.

**Graphical Abstract:**

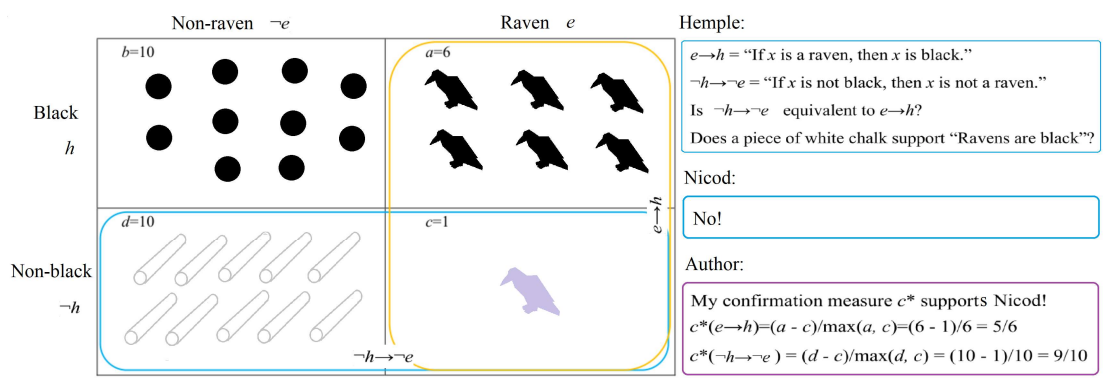

## 1. Introduction

A universal judgment is equivalent to a hypothetical judgment or a rule, such as "All ravens are black" is equivalent to "For every $x$, if $x$ is a raven, then $x$ is black". Both can be used as a major premise for a syllogism. Deductive logic needs major premises; however, some major premises for empirical reasoning must be supported by inductive logic. Logical empiricism affirmed that a universal judgment can be verified finally by sense data. Popper said against logical empiricism that a universal judgment could only be falsified rather than be verified. After long arguments, Popper and most logical empiricists reached the identical conclusion that uncertain universal judgments or major premises could be confirmed by evidence [1,2].

In 1945, Hemple [3] discovered the confirmation paradox or the Raven Paradox (RP). According to the Equivalence Condition (EC) in the classical logic, "If $x$ is a raven, then $x$ is black" (Rule I) is equivalent to "If $x$ is not black, then $x$ is not a raven" (Rule II). A piece of white chalk supports the Rule II, and hence, also supports the Rule I. However, according to the Nicod Criterion (NC) [4] or Nicod-Fisher Criterion (NFC) (see Section 2.3 for details), a black raven supports the Rule I, a non-black raven undermines the Rule I, and a non-raven thing, such as a black cat or a piece of white chalk, is irrelevant to the Rule I. Hence, there exists a paradox between EC and NC or NFC.

To quantize confirmation, both Carnap [1] and Popper [2] proposed their confirmation measures. However, only Carnap's confirmation measures are famous. So far, researchers have proposed many confirmation measures [1,5-13]. The induction problem seemly have become the confirmation problem. To screen reasonable confirmation measures, Elles and Fitelson [14] proposed symmetries and asymmetries as desirable properties; Crupi *et al.* [8] and Greco *et al.* [15] suggested normalization (for measures between -1 and 1) as a desirable property; Greco *et al.* [16] proposed monotonicity as a desirable property. We can find that only measures $F$ (proposed by Kemeny and Oppenheim) and $Z$ among popular confirmation measures possess these desirable properties. Measure $Z$ was proposed by Crupi *et al.* [8] as the normalization of some other confirmation measures. It is also called the certainty factor proposed by Shortliffe and Buchanan [7].

When the author of this paper [17] optimized the (fuzzy) truth functions using his semantic information method, he found that any truth function for binary classifications could be treated as the combination of a clear prediction's truth function (with a value 0 or 1) and a tautology's truth function (with a constant value 1). The combining proportion of the former optimized by a sampling distribution may be regarded as the confirmation measure. This measure is denoted by $b^*$; it is similar to measure $F$ and also possesses the above-mentioned desirable properties.

In the medical society, Likelihood Ratio (LR) [18] is used to assess the quality of medical tests; positive $LR$, e.g., $LR+$, is used to assess the uncertain major premise "If $x$ is positive, then $x$ is infected", where $x$ is a person's specimen or datum. Measures $L$, $F$, and $b^*$ are the functions of $LR$. $LR$ changes between 0 and $\infty$; $L$ changes between $-\infty$ and $\infty$. Both $LR$ and $L$ cannot indicate the difference between a test and the best test (or the worst test) as well as $F$ and $b^*$. However, it is still not easy to use $F$, $b^*$, or other measures to eliminate the RP.

Recently, the author found that the problem with the RP is different from the problem with the medical test. $LR$, $F$, and $b^*$ indicate how good the testing means are instead of how good the probability predictions are. To clarify the RP, we need a confirmation measure that can indicate how good the probability predictions are. The confirmation measure $c^*$ is hence derived. We call $c^*$ a prediction confirmation measure and call $b^*$ a channel confirmation measure. The distinction between Channels' confirmation and predictions' confirmation is similar to yet different from the distinction between Bayesian confirmation and Likelihoodist confirmation [19]. Measure $c^*$ accords with the NFC and undermines the EC, and hence can be used to eliminate the RP.

The main purposes of this paper are 1) to use a confirmation measure to eliminate the RP, and 2) to explain that confirmation and falsification may be compatible.

The confirmation methods in this paper are different from popular methods since

● $b^*$ and $c^*$ are derived by the semantic information method [17, 20] and the maximum likelihood criterion rather than defined.

● confirmation and statistical learning mutually support so that the confirmation measures are used not only to assess major premises but also to make probability predictions.

The main contributions of this paper are

● It distinguishes two types of confirmation measures: 1) the channel confirmation measure, which is similar to $LR$ and indicates how good a channel is and 2) the prediction confirmation measure, which is similar to accuracy (or the correct rate) and indicates how good a probability prediction is.

● It provides measure $c^*$ that manifests NFC, and hence can be used to eliminate the RP.

The rest of this paper is organized as follows. Section 2 includes methods. It reviews existing confirmation measures and their screening; it also introduces how to use the semantic information method to derive new confirmation measures $b^*$ and $c^*$ with the medical test as an example. Section 3 includes results. It provides equations with four examples' numbers *a, b, c,* and *d* to express new confirmation measures for a major premise with different antecedents and seccedents. It also provides some equations to explain relationships between $F$ and $b*$ and some examples to show the characteristics of new confirmation measures. Section 4 discusses how we can eliminate the RP by measure $c*$. It also discusses some conceptual confusions and misunderstandings in the popular confirmation methods and explains how new confirmation measures are compatible with Popper's falsification thought. Section 5 ends with conclusions.

## 2. Methods

### 2.1. To Review popular confirmation measures

We use $h_1$ to denote a hypothesis, $h_0$ to denote its negation, and $h$ to denote one of them. We use $e_1$ as another hypothesis treated as the evidence of $h_1$, $e_0$ as its negation, and $e$ as one of them. We use $c(e, h)$ to represent a confirmation measure, which means the degree of inductive support. Note that $c(e, h)$ here is used as in [8], where $e$ is on the left, and $h$ is on the right.

The Popular confirmation measures include：

$D(e_1, h_1) = P(h_1|e_1) - P(h_1)$ (Carnap, 1962 [1]),

$M(e_1, h_1) = P(e_1|h_1) - P(e_1)$ (Mortimer, 1988 [5]),

$R(e_1, h_1) = \log[P(h_1|e_1)/P(h_1)]$ (Horwich, 1982 [6]),

$C(e_1, h_1) = P(h_1, e_1) - P(e_1) P(h_1)$ (Carnap, 1962 [1]),

$Z(h_1, e_1) = \begin{cases} [P(h_1 \mid e_1) - P(h_1)]/[1 - P(h_1)], & \text{as } P(h_1 \mid e_1) \geq P(h_1), \\ [P(h_1 \mid e_1) - P(h_1)]/P(h_1), & \text{otherwise}, \end{cases}$ (Shortliffe and Buchanan, 1975 [7], Crupi $et\ al.$, 2007 [8]),

$S(e_1, h_1) = P(h_1|e_1) - P(h_1|e_0)$ (Christensen, 1999 [9]),

$N(e_1, h_1) = P(e_1|h_1) - P(e_1|h_0)$ (Nozik, 1981 [10]),

$L(e_1, h_1) = \log[ P(e_1|h_1)/P(e_1|h_0)]$ (Good, 1984 [11]), and

$F(e_1, h_1) = [ P(e_1|h_1) - P(e_1|h_0)]/[ P(e_1|h_1) + P(e_1|h_0)]$ (Kemeny and Oppenheim, 1952 [12]).

There are more confirmation measures in [8,21]. $F$ is also called $l*$ [13], $L$ [8], or $k$ [21].

Firstly, we need to clarify that confirmation is to assess what kind of evidence supports what kind of hypotheses. Let us have a look at the following three hypotheses:

- **Hypothesis 1**: $h_1(x)$="$x$ is elderly." where $x$ is a variable for an age and $h_1(x)$ is a predicate. An instance $x$=70 may be the evidence, and the truth value of proposition $h_1(70)$ should be 1. If $x$=50, the (fuzzy) truth value should be less, such as 0.5. Let $e_1$="$x{\geq}60$", true $e_1$ may also be the evidence that supports $h_1$ so that $P(h_1|e_1) > P(h_1)$, where $P$ means logical probability.

- **Hypothesis 2**: $h_1(x)$="If age $x{\geq}60$, then $x$ is elderly", which is a hypothetical judgment, a major premise, or a rule. Note that $x$=70 or $x{\geq}60$ is only the evidence of the consequence "$x$ is elderly" instead of the evidence of the rule. The rule's evidence should be a sample with many examples.

- **Hypothesis 3**: $e_1 \rightarrow h_1$="If age $x{\geq}60$, then $x$ is elderly", which is the same as Hypothesis 2. The difference is that $e_1$="$x{\geq}60$"; $h_1$="$x$ is elderly". The evidence is a sample with many examples like $\{(e_1, h_1), (e_1, h_0),...\}$, or a sampling distribution $P(e, h)$, where $P$ means statistical probability.

**Hypothesis 1** only has a truth function or logical probability between 0 and 1, which is ascertained by our definition or usage. **Hypothesis 1** does not need to be confirmed. **Hypothesis 2** or **Hypothesis 3** is what we need to confirm. The degree of confirmation is between -1 and 1.

There exist two different understandings about $c(e, h)$:

- **Understanding 1**: The $h$ is the major premise to be confirmed, and $e$ is the evidence that supports $h$; $h$ and $e$ are so used by Elles and Fitelson [14].

- **Understanding 2**: The $e$ and $h$ are those in rule $e{\rightarrow}h$ as used by Kemeny and Oppenheim [12]. The $e$ is only the evidence that supports the conclusion $h$ instead of the major premise $e{\rightarrow}h$.

Fortunately, although researchers understand $c(e, h)$ in different ways, most researchers agree to use a sample including four types of examples $(e_1, h_1)$, $(e_0, h_1)$, $(e_1, h_0)$, and $(e_0, h_0)$ as the evidence to confirm a rule and to use the four examples' numbers $a$, $b$, $c$, and $d$ (see Table 1) to construct confirmation measures. The following statements are based on this common view.

Table 1. The numbers of four types of examples for confirmation measures

|  | $e_0$ | $e_1$ |
|---|---|---|
| $h_1$ | $b$ | $a$ |
| $h_0$ | $d$ | $c$ |

The $a$ is the number of the example $(e_1, h_1)$. For example, $e_1=$"raven" ("raven" is a label or the abbreviate of "$x$ is a raven") and $h_1=$"black"; $a$ is the number of black ravens. Similarly, $b$ is the number of black non-raven things; $c$ is the number of non-black ravens; $d$ is the number of non-black and non-raven things.

To make the confirmation task clearer, we follow **Understanding 2** to treat $e{\rightarrow}h$ as "if $e$ then $h$" as the rule to be confirmed and replace $c(e, h)$ with $c(e{\rightarrow}h)$. To research confirmation is to construct or select the function $c(e{\rightarrow}h)=f(a, b, c, d)$.

To screen reasonable confirmation measures, Elles and Fitelson [14] proposed the following symmetries:

- Hypothesis Symmetry (HS): $c(e_1{\rightarrow}h_1)=-c(e_1{\rightarrow}h_0)$ (two seccedents are opposite),
- Evidence Symmetry (ES): $c(e_1{\rightarrow}h_1)=-c(e_0{\rightarrow}h_1)$ (two antecedents are opposite),
- Commutativity Symmetry (CS): $c(e_1{\rightarrow}h_1)=c(h_1{\rightarrow}e_1)$, and
- Total Symmetry (TS): $c(e_1{\rightarrow}h_1)=c(e_0{\rightarrow}h_0)$.

They conclude that only HS is desirable; the other three symmetries are not desirable. We call this conclusion as the Symmetry/Asymmetry requirement. Their conclusion is supported by most researchers. Since TS is the combination of HS and ES, we only need to check HS, ES, and CS. According to this symmetry and asymmetry requirement, only $L$, $F$, and $Z$ among the measures mentioned above are screened out. It is uncertain whether $N$ can be ruled out by this requirement [15]. See [14,22,23] for more discussions about the symmetries.

Greco *et al.* [15] proposed monotonicity as a desirable property. If $f(a, b, c, d)$ does not decrease with $a$ or $d$ and does not increase with $b$ or $c$, then we say that $f(a, b, c, d)$ has the monotonicity. $L$, $F$, and $Z$ have this monotonicity, whereas $D$, $M$, $N$ do not have. If we further require that $c(e{\rightarrow}h)$ are normalizing (between -1 and 1) [8, 12], then only $F$ and $Z$ are screened out. There are also other properties are discussed [15,19]. One is Logicality, which means $c(e{\rightarrow}h)=1$ without counterexample and $c(e{\rightarrow}h)=-1$ without positive example. We can also screen out $F$ and $Z$ using the Logicality requirement. Another is Maximality/Minimality, which means $c(e_1{\rightarrow}h_1)$ should be maximal iff $c=b=0$ and minimal iff $a=d=0$. However, $F$ does not satisfy Maximality/Minimality requirement [21]. The reason is that $b=0$ is not necessary to make $F(e_1{\rightarrow}h_1)=1$, and $c=0$ is not necessary to make $F(h_1{\rightarrow}e_1)=1$. It is easy to find that Maximality/Minimality is compatible with Commutativity Symmetry.

Some researchers pay attention to the Maximality/Minimality requirement for different applications. For example, Glass [21] proposed two confirmation measures of association rule interestingness, which satisfies Maximality/Minimality requirement and dissatisfies the Symmetry/Asymmetry requirement. The author of this paper pays attention to the Symmetry/Asymmetry requirement instead of the Maximality/Minimality requirement for confirming major premises.

Consider the medical test, such as the test for HIV (Human Immunodeficiency Virus). Let $e_1=$"positive" (e.g., "$x$ is positive", where $x$ is a specimen), $e_0=$"negative", $h_1=$"infected" (e.g., "$x$ is infected"), and $h_0$="uninfected". Then the positive likelihood ratio is $LR^+ = P(e_1|h_1)/P(e_1|h_0)$, which indicates the reliability of the rule $e_1 \rightarrow h_1$. $L$ and $F$ have the one-to-one correspondence with $LR$:

$$L(e_1 \rightarrow h_1) = \log LR^+; \tag{1}$$

$$F(e_1, h_1) = (LR^+ - 1)/(LR^+ + 1). \tag{2}$$

Hence, $L$ and $F$ can also be used to assess the reliability of the medical test. In comparison with $LR$ and $L$, $F$ indicates the distance between the test ($F$) and the best test (1) or the worst test (-1). However, $LR$ can be used for the probability predictions of diseases more conveniently [18].

### 2.2. Shannon's channel and Semantic channel of medical tests

Consider the medical test. The relationship between $h$ and $e$ is shown in Figure 1.

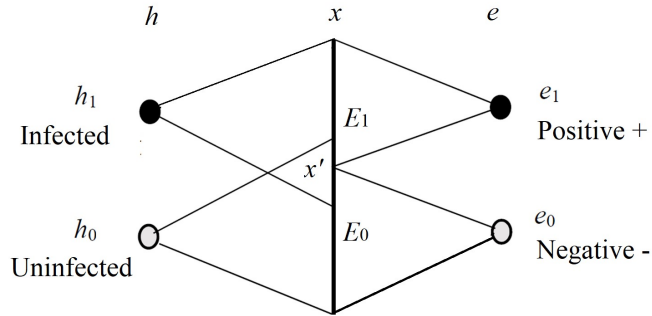

Figure 1. The relationship between Positive/Negative and Infected/Uninfected in the medical test

The $x \in U$ denotes a specimen or a testing datum; $x'$ is the partitioning boundary of testing data. Classifications for elder people, good watermelons, and junk emails are similar. When we classify people into elderly people and non-elderly people, $x$ may be any age, $x'$ may be age 60. The $h_1$ is the true label "elderly" in a sample, and $e_1$ is the classified label "elderly" according to $x$. To optimize classifications, we may use the maximum posterior probability criterion, which is equivalent to the maximum correctness criterion, or the maximum likelihood ratio criterion. The author has proved that the maximum likelihood ratio criterion is equivalent to the maximum mutual information criterion [24]; we may use an iterative algorithm to achieve the maximum mutual information classifications [17].

Figure 2 shows the relationship between $h$ and $x$ by two posterior probability distributions $P(x|h_0)$ and $P(x|h_1)$ and the relationship between $e$ and $x$ by two crisp sets $E_0$ and $E_1$.

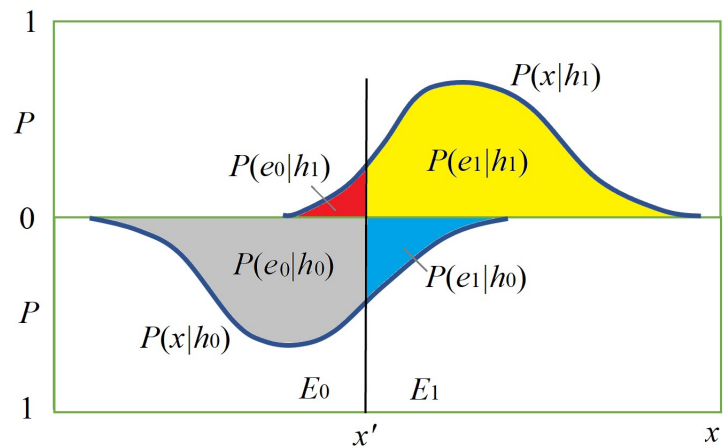

Figure 2. Four conditional probabilities for $LR^+$, $LR^-$, and Shannon's channel $P(e|h)$

The two areas that the two function curves cover may be treated as the elements of two fuzzy sets $\theta_1$ and $\theta_0$ [17]. Let $E_1 \rightarrow \theta_1$ ="If $x$ is in $E_1$, then $x$ is in $\theta_1$". Then we can regard $E_1 \rightarrow \theta_1$ as $e_1 \rightarrow h_1$ for intuitionistic understanding.

In the medical test, $P(e_1|h_1)$ is called sensitivity, and $P(h_0|e_0)$ is called specificity. They ascertain a Shannon channel [25], as shown in Table 2.

Table 2. The sensitivity and specificity ascertain a Shannon's Channel $P(e|h)$

|  | Negative $e_0$ | Positive $e_1$ |
|---|---|---|
| Infected $h_1$ | $P(e_0|h_1)$=1-sensitivity | $P(e_1|h_1)$=sensitivity |
| Uninfected $h_0$ | $P(e_0|h_0)$=specificity | $P(e_1|h_0)$=1-specificity |

Next, we define the semantic channel [17]. Let $e_1(h)$ be a predicate. Its truth function is $T(\theta_{e1}|h)$, where $\theta_{e1}$ a fuzzy subset on $\{h_0, h_1\}$. We may regard $e_1(h)$ as the combination of believable and disbelievable parts (see Figure 3). The truth function of the believable part is $T(E_1|h) \in \{0,1\}$. The disbelievable part is a tautology, whose truth function is always 1. Then we have truth functions of predicates $e_1(h)$ and $e_0(h)$:

$$T(\theta_{e1}|h)= b_1' +b_1'T(E_1|h); \qquad T(\theta_{e0}|h)= b_0' +b_0'T(E_0|h). \qquad (3)$$

where $b_1'$ is the proportion of the disbelievable part and also the truth value $T(\theta_{e1}|h_0)$ of $e_1$ for given counter-instance $h_0$ as the condition.

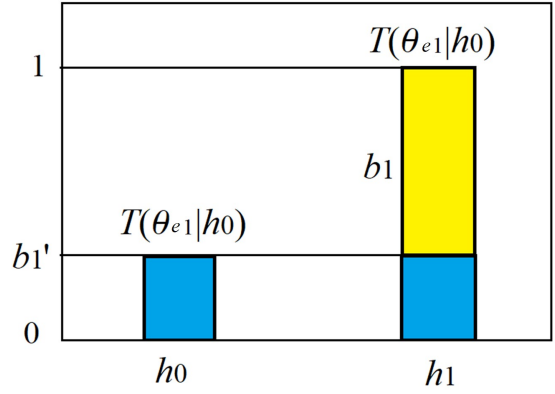

Figure 3. Truth function $T(\theta_{e1}|h)$ is regarded as the believable part ($b_1$) plus the disbelievable part ($b_1'$=1-$|b_1|$).

The four truth values form a semantic channel, as shown in Table 3.

Table 3. The semantic channel ascertained by $b_1'$ and $b_0'$ for the medical test

|  | $e_0$ (negative) | $e_1$ (positive) |
|---|---|---|
| $h_1$ (infected) | $T(\theta_{e0}|h_1)=b_0'$ | $T(\theta_{e1}|h_1)=1$ |
| $h_0$ (uninfected) | $T(\theta_{e0}|h_0)=1$ | $T(\theta_{e1}|h_0)=b_1'$ |

### 2.3. Semantic information formulas and the Nicod-Fisher criterion

We can obtain the posterior probability distribution $P(h|\theta_{e1})$ of $h$ from $P(h)$ and $T(\theta_{e1}|h)$ by the semantic Bayes' formula [17]

$$P(h \mid \theta_{e1}) = P(h)T(\theta_{e1} \mid h) / T(\theta_{e1}),$$
$$T(\theta_{e1}) = \sum_i P(h_i)T(\theta_{e1} \mid h_i) = P(h_1) + b_1'P(h_0), \qquad (4)$$

where $T(\theta_{e1})$ is the logical probability of predicate $e_1(h)$. The distinction and relationship between logical probability and statistical probability are discussed in [17].

According to the semantic information G theory [17,20], the semantic information conveyed by $e_1$ about $h$ is

$$I(h; \theta_{e1}) = \log \frac{P(h_i | \theta_{e1})}{P(h_i)} = \log \frac{T(\theta_{e1} | h)}{T(\theta_{e1})}. \tag{5}$$

The average semantic information is

$$I(H; \theta_{e1}) = \sum_{i=0}^{1} P(h_i | e_1) \log \frac{T(\theta_{e1} | h_i)}{T(\theta_{e1})} = \sum_{i=0}^{1} P(h_i | e_1) \log \frac{P(h_i | \theta_{e1})}{P(h_i)}, \tag{6}$$

where $H$ is a random variable taking value $h$; $P(h_i | e_1)$ is the conditional probability from the sample.

We now consider the relationship between the likelihood and the average semantic information.

Let **D** be a sample $\{(h(t), e(t)) | t = 1$ to $N; h(t) \in \{h_0, h_1\}; e(t) \in \{e_0, e_1\}\}$, which includes two sub-samples or conditional samples **H₀** with label $e_0$ and **H₁** with label $e_1$. When $N$ data points in **D** come from Independent and Identically Distributed (IID) random variables, we have the log-likelihood

$$L(\theta_{e1}) = \log P(\mathbf{H}_1 | \theta_{e1}) = \log P(h(1), h(2), ..., h(N) | \theta_{e1}) = \log \prod_{i=0}^{1} P(h_i | \theta_{e1})^{N_{1i}}$$

$$= N_1 \sum_{i=0}^{1} P(h_i | e_1) \log P(h_i | \theta_{ej}) = -N_1 H(H | \theta_{e1}). \tag{7}$$

where $\theta_{e1}$ is a fuzzy set and can also be treated as a model or a model's parameters; $N_{1i}$ is the number of the example $(h_i, e_1)$ in **D**; $N_1$ is the size of **H₁**. Comparing the above two equations, we have

$$I(H; \theta_{e1}) = L(\theta_{e1}) / N_1 - \sum_{i=0}^{1} P(h_i | e_1) \log P(h_1). \tag{8}$$

Since the second term on the right side is constant, the maximum likelihood criterion is equivalent to the maximum average semantic information criterion. It is easy to find that a positive example $(e_1, h_1)$ increases the average log-likelihood $L(\theta_{e1})/N_1$; a counterexample $(e_0, h_1)$ decreases it; examples $(e_0, h_0)$ and $(e_0, h_1)$ with $e_0$ are irrelevant to it.

The Nicod criterion about confirmation is that a positive example $(e_1, h_1)$ supports rule $e_1 \rightarrow h_1$; a counterexample $(e_1, h_0)$ undermines $e_1 \rightarrow h_1$. No reference exactly indicates if Nicod affirmed that $(e_0, h_1)$ and $(e_0, h_1)$ are irrelevant to $e_1 \rightarrow h_1$. If Nicod did not affirm, we can add this affirmation to the criterion. We can call the corresponding criterion the Nicod-Fisher criterion, since Fisher proposed Maximum Likelihood (ML) estimation.

## 3. To derive two confirmation measures b* and c*

### 3.1. To derive channel confirmation measure b*

Logical Bayesian Inference (LBI) based on the semantic information G theory provides the method of optimizing truth functions by sampling distributions [17]. It concludes that as $P(h|\theta_{e1})=P(h|e_1)$, the average semantic information $I(H; \theta_{e1})$ reaches its maximum. Meanwhile, the truth function $T*(\theta_{e1}|h)$ is proportional to the transition probability function [25], e.g., $T*(\theta_{e1}|h) \propto P(e_1|h)$. Hence

$$\frac{1}{P(e_1 | h_1)} = \frac{b_1{}^{'*}}{P(e_1 | h_0)}. \tag{9}$$

where $b_1{}^{'*}$ is the optimized degree of disbelief and is called the degree of disconfirmation of rule $e_1 \rightarrow h_1$. From the above equation, we can obtain

$$b_1{}^{'*} = P(e_1|h_0) / P(e_1|h_1) = 1/LR^+, \tag{10}$$

where $P(h_1|e_1) \geq P(h_0|e_1)$. We call

$$b_1* = 1 - b_1{}^{'*} = [P(e_1|h_1) - P(e_1|h_0)]/P(e_1|h_1) \tag{11}$$

the degree of confirmation of the rule $e_1 \to h_1$. Consider $P(h_1|e_1) < P(h_0|e_1)$. We have

$$b_1^* = b_1'^* - 1 = [P(e_1|h_0) - P(e_1|h_1)]/P(e_1|h_0). \quad (12)$$

Combining the above two formulas, we obtain

$$b_1^* = b^*(e_1 \to h_1) = \frac{P(e_1|h_1) - P(e_1|h_0)}{\max[P(e_1|h_1), P(e_1|h_0)]} = \frac{LR^+ - 1}{\max(LR^+, 1)}. \quad (13)$$

In the same way, we obtain

$$b_0^* = b^*(e_0 \to h_0) = \frac{P(e_0|h_0) - P(e_0|h_1)}{\max[P(e_0|h_0), P(e_0|h_1)]} = \frac{LR^- - 1}{\max(LR^-, 1)}. \quad (14)$$

where $LR^- = P(h_0|e_0)/P(h_1|e_0)$ is the negative likelihood ratio. Using HS, we can obtain $b^*(e_1 \to h_0) = -b^*(e_1 \to h_1)$ and $b^*(e_0 \to h_1) = -b^*(e_0 \to h_0)$.

Compared with $F$, $b^*$ is better for probability predictions. For example, from $b_1^* > 0$ and $P(h)$, we obtain

$$P(h_1|\theta_{e1}) = P(h_1)/[P(h_1) + b_1'^* P(h_0)] = P(h_1)/[1 - b_1^* P(h_0)]. \quad (15)$$

If $b_1^* = 0$, then $P(h_1|\theta_{e1}) = P(h_1)$. If $P(h_1|\theta_{e1}) < 0$, then we can make use of HS to have $b_0^* = b_1^*(e_1 \to h_0) = |b_1^*(e_1 \to h_1)|$. Then we obtain

$$P(h_0|\theta_{e1}) = P(h_0)/[P(h_0) + b_0'^* P(h_1)] = P(h_0)/[1 - b_0^* P(h_1)]. \quad (16)$$

We can also obtain $b_1^* = 2F_1/(1 + F_1)$ from $F_1 = F(e_1 \to h_1)$ for $P(h_1|\theta_{e1})$, but the calculation of probability predictions with $F_1$ is a little complicated.

So far, it is still problematic to use $b^*$, $F$, or another measure to handle the RP. For example, the increments of $b^*(e_1 \to h_1)$ caused by $d$ plus 1 and $a$ plus 1 may be similar, which means that a piece of white chalk can support "Ravens are black" as well as a black raven. Hence $b^*$ and $F$ do not accord with NFC. $Z$ does not too.

Why does not measure $b^*$ and $F$ accord with NFC? The reason is that the likelihood $L(\theta_{e1})$ is related to $P(h)$, whereas $b^*$ and $F$ are irrelevant to $P(h)$.

### 3.2. To derive prediction confirmation measure $c^*$

Statistics not only uses Likelihood Ratio (LR) to indicate how reliable a testing means (as a channel) is but also uses the Correct Rate (CR) to indicate how reliable a probability prediction is. Like $LR$, $F$ and $b^*$ cannot indicate the quality of a probability prediction. Most other measures have similar problems as mentioned by [9].

For example, a testing method of HIV [26] has sensitivity $P(e_1|h_1) = 0.917$ and specificity $P(e_0|h_0) = 0.001$. We can calculate $b_1^* = 0.9989$. When the prior probability of $h_1$ changes, predicted probability $P(h_1|\theta_{e1})$ according to Eq. (15), which is also called positive predictive value, changes with the prior probability of $h_1$, as shown in Table 4. We can obtain the same results using the classical Bayes' method.

Table 4. Predictive probability $P(h_1|\theta_{e1})$ changes with prior probability $P(h_1)$ as $b_1^* = 0.9989$.

|                        | Safe group | Common people | High-risky group |
|------------------------|-----------|---------------|------------------|
| $P(h_1)$               | 0.0001    | 0.002         | 0.1              |
| $P(h_1|\theta_{e1})$   | 0.0084    | 0.65          | 0.99             |

Data in Table 4 shows that $b^*$ or $F$ cannot indicate the quality of probability predictions. Therefore, we need to use $P(h)$ to construct a confirmation measure that is similar to the CR.

We now treat $P(h|\theta_{e1})$ as the combination of a useful part with proportion $c_1$ and a useless part with proportion $c_1$', as shown in Figure 4. We call $c_1^*$ the degree of belief of $e_1 \to h_1$ as a prediction.

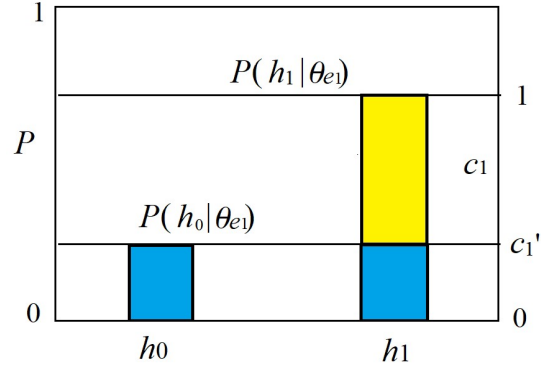

Figure 4. Likelihood function $P(h|\theta_{e1})$ may be regarded as a useful part plus a useless part.

As the prediction accords with the fact, e.g., $P(h|\theta e_1)= P(h|e_1)$, $c_1$ becomes $c_1{}^*$. The degree of disconfirmation for predictions is

$$c'{}^*(e_1 \rightarrow h_1)=\min(\ P(h_0|e_1)/P(h_1|e_1),\ P(h_1|e_1)/P(h_0|e_1)). \quad (17)$$

Further, we have

$$\begin{aligned} c*(e_1 \rightarrow h_1) &= \frac{P(h_1 \mid e_1) - P(h_0 \mid e_1)}{\max(P(h_1 \mid e_1), P(h_0 \mid e_1))} \\ &= \frac{2P(h_1 \mid e_1) - 1}{\max(P(h_1 \mid e_1), 1 - P(h_1 \mid e_1))} = \frac{2CR_1 - 1}{\max(CR_1, 1 - CR_1)}. \end{aligned} \quad (18)$$

where $CR_1=P(h_1|\theta_{e1})=P(h_1|e_1)$ is the correct rate of rule $e_1 \rightarrow h_1$. This correct rate means that the probability of $H=h_1$ as $x \in E_1$ is $CR_1$. Letting both the numerator and denominator of Eq. (18) be multiplied by $P(e_1)$, we obtain

$$c_1* = c*(e_1 \rightarrow h_1) = \frac{P(h_1, e_1) - P(h_0, e_1)}{\max(P(h_1, e_1), P(h_0, e_1))} = \frac{a - c}{\max(a, c)}. \quad (19)$$

The sizes of four areas covered by two curves in Figure 5 are proportional to $a$, $b$, $c$, and $d$.

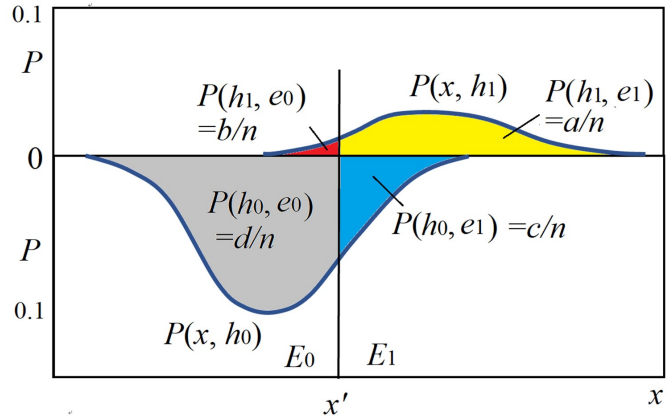

Figure 5. The numbers of positive examples and counterexamples for $c*(e_0 \rightarrow h_0)$ (see the left side) and $c*(e_1 \rightarrow h_1)$ (see the right side).

Using the similar method, we obtain

$$c_0* = c*(e_0 \rightarrow h_0) = \frac{P(h_0, e_0) - P(h_1, e_0)}{\max(P(h_0, e_0), P(h_1, e_0))} = \frac{d - b}{\max(d, b)}. \quad (20)$$

Making use of HS or Seccedent Symmetry (SS), we can obtain $c^*(e_1 \to h_0) = -c^*(e_1 \to h_1)$ and $c^*(e_0 \to h_1) = -c^*(e_0 \to h_0)$.

In Figure 5, the sizes of the two areas covered by two curves are $P(h_0)$ and $P(h_1)$, which are different. If $P(h_0)=P(h_1)=0.5$, then the prediction confirmation measure $c^*$ is equal to the channel confirmation measure $b^*$.

We can also use $c^*$ for probability predictions. As $c_1^*>0$, according to Eq. (18), we have

$$P(h_1 \mid \theta_{e1}) = CR_1 = 1/(1+c_1'^*) = 1/(2-c_1^*). \tag{21}$$

If $c^*(e_1 \to h_1)=0$ then $P(h_1|\theta_{e1})=0.5$. If $c^*(e_1 \to h_1)<0$, we may make use of HS or SS to have $c_0^*=c^*(e_1 \to h_0)=|c^*(e_1 \to h_1)|$, and then make probability prediction:

$$P(h_0 \mid \theta_{e1}) = 1/(2-c_0^*),$$
$$P(h_1 \mid \theta_{e1}) = 1 - P(h_0 \mid \theta_{e1}) = (1-c_0^*)/(2-c_0^*). \tag{22}$$

We may define another prediction confirmation measure by replacing "max( )" with "+":

$$c_{F1} = c_F *(e_1 \to h_1) = \frac{P(h_1|e_1) - P(h_0|e_1)}{P(h_1|e_1) + P(h_0|e_1)} = P(h_1|e_1) - P(h_0|e_1)$$
$$= \frac{P(h_1, e_1) - P(h_0, e_1)}{P(e_1)} = \frac{a-c}{a+c}. \tag{23}$$

The $c_F^*$ is convenient for probability predictions as $P(h)$ is certain. There is

$$P(h_1 \mid \theta_{e1}) = CR_1 = (1+c_{F1}^*)/2;$$
$$P(h_0 \mid \theta_{e1}) = 1 - CR_1 = (1-c_{F1}^*)/2. \tag{24}$$

However, when $P(h)$ is variable, we should use $b^*$ with $P(h)$ for probability predictions.

It is easy to prove that $c^*(e_1 \to h_1)$ and $c_F^*(e_1 \to h_1)$ possess all the above-mentioned desirable properties.

### 3.3. Converse channel/prediction confirmation measures $b^*(h \to e)$ and $c^*(h \to e)$

Greco et al. [19] divide confirmation measures into
- Bayesian confirmation measures with $P(h|e)$ for $e \to h$,
- Likelihoodist confirmation measures with $P(e|h)$ for $e \to h$,
- converse Bayesian confirmation measures with $P(h|e)$ for $h \to e$, and
- converse Likelihoodist confirmation measures with $P(e|h)$ for $h \to e$.

Similarly, this paper divides confirmation measures into
- channel confirmation measure $b^*(e \to h)$,
- prediction confirmation measure $c^*(e \to h)$,
- converse channel confirmation measure $b^*(h \to e)$, and
- converse prediction confirmation measure $c^*(h \to e)$.

We now consider $c^*(h_1 \to e_1)$. Suppose that $P(x|h_1)$ is the age distribution of people called "elderly". If one knows $P(x|h_1)$ and is asked to explain the meaning of $P(x|h_1)$ in natural language, he may explain "If one is called elderly, maybe he/she is over 60 years old". This kind of explanation can be abstracted as $h_1 \to e_1$. The confirmation measure $c^*(h_1 \to e_1)$ is a converse prediction confirmation measure. The positive examples' proportion and counterexamples' proportion can be found in the upside of Figure 5. Then we have

$$c^*(h_1 \to e_1) = \frac{P(e_1 \mid h_1) - P(e_0 \mid h_1)}{\max(P(e_1 \mid h_1), P(e_0 \mid h_1))} = \frac{a-b}{\max(a,b)}. \tag{25}$$

The correct rate reflected by $c^*(h_1 \to e_1)$ is sensitivity or true positive rate $P(h_1|e_1)$. The correct rate reflected by $c^*(h_0 \to e_0)$ is specificity or true negative rate $P(h_0|e_0)$.

Consider the converse channel confirmation measure $b^*(h_1 \to e_1)$. Now the source is $P(e)$ instead of $P(h)$. We may swap $e_1$ with $h_1$ in $b^*(e_1 \to h_1)$ or $a$ with $d$ in $f(a, b, c, d)$ to obtain

$$b^*(h_1 \to e_1) = \frac{P(h_1 \mid e_1) - P(h_1 \mid e_0)}{P(h_1 \mid e_1) \vee P(h_1 \mid e_0)} = \frac{ad - bc}{a(b+d) \vee b(a+c)}. \tag{26}$$

where $\vee$ is the max operator in fuzzy logic and is used to replace max( ). There are also four types of converse channel/prediction confirmation formulas with $a$, $b$, $c$, and $d$ (see Table 8). Due to HS or SS, there are eight types of converse channel/prediction confirmation formulas altogether.

### 3.4. Eight confirmation formulas for different antecedents and seccedents

Table 5 and Table 6 shown the positive examples' and counterexamples' proportions needed by $b^*$ and $c^*$.

Table 5. Four proportions for calculating $b^*(e \to h)$ and $c^*(e \to h)$

|  | $e_0$ (negative) | $e_1$ (positive) |
|---|---|---|
| $h_1$ (infected) | $P(e_0\|h_1)=b/(a+b)$ | $P(e_1\|h_1)=a/(a+b)$ |
| $h_0$ (uninfected) | $P(e_0\|h_0)=d/(c+d)$ | $P(e_1\|h_0)=c/(c+d)$ |

Table 6. Four proportions for calculating $b^*(h \to e)$和 $c^*(h \to e)$

|  | $e_0$ (negative) | $e_1$ (positive) |
|---|---|---|
| $h_1$ (infected) | $P(h_1\|e_0)=b/(b+d)$ | $P(h_1\|e_1)=a/(a+c)$ |
| $h_0$ (uninfected) | $P(h_0\|e_0)=d/(b+d)$ | $P(h_0\|e_1)=c/(a+c)$ |

Table 7 provides four types of confirmation formulas with $a$, $b$, $c$, and $d$ for rule $e \to h$, where function max() is replaced with the operator $\vee$. These confirmation measures are related to the false report rates of the rule $e \to h$.

Table 7. Channel/prediction confirmation measures expressed by $a$, $b$, $c$, and $d$.

|  | $b^*(e \to h)$ (for channels, refer to Figure 2) | $c^*(e \to h)$ (for predictions, refer to Figure 5) |
|---|---|---|
| $e_1 \to h_1$ | $\dfrac{P(e_1 \mid h_1) - P(e_1 \mid h_0)}{P(e_1 \mid h_1) \vee P(e_1 \mid h_0)} = \dfrac{ad - bc}{a(c+d) \vee c(a+b)}$ | $\dfrac{P(h_1\|e_1) - P(h_0\|e_1)}{P(h_1\|e_1) \vee P(h_0\|e_1)} = \dfrac{a-c}{a \vee c}$ |
| $e_0 \to h_0$ | $\dfrac{P(e_0 \mid h_0) - P(e_0 \mid h_1)}{P(e_0 \mid h_0) \vee P(e_0 \mid h_1)} = \dfrac{ad - bc}{d(a+b) \vee b(c+d)}$ | $\dfrac{P(h_0 \mid e_0) - P(h_1 \mid e_0)}{P(h_0 \mid e_0) \vee P(h_1 \mid e_0)} = \dfrac{d-b}{d \vee b}$ |

Table 8 includes four types of confirmation measures for $h \to e$. These confirmation measures are related to the missing report rates of the rule $h \to e$.

Table 8. Converse channel/prediction confirmation measures expressed by $a$, $b$, $c$, and $d$.

|  | $b^*(h \to e)$ (for converse channels) | $c^*(h \to e)$ (for converse predictions, refer to Figure 5) |
|---|---|---|
| $h_1 \to e_1$ | $\dfrac{P(h_1 \mid e_1) - P(h_1 \mid e_0)}{P(h_1 \mid e_1) \vee P(h_1 \mid e_0)} = \dfrac{ad - bc}{a(b+d) \vee b(a+c)}$ | $\dfrac{P(e_1 \mid h_1) - P(e_0 \mid h_1)}{P(e_1 \mid h_1) \vee P(e_0 \mid h_1)} = \dfrac{a-b}{a \vee b}$ |
| $h_0 \to e_0$ | $\dfrac{P(h_0 \mid e_0) - P(h_0 \mid e_1)}{P(h_0 \mid e_0) \vee P(h_0 \mid e_1)} = \dfrac{ad - bc}{d(a+c) \vee c(b+d)}$ | $\dfrac{P(e_0 \mid h_0) - P(e_1 \mid h_0)}{P(e_0 \mid h_0) \vee P(e_1 \mid h_0)} = \dfrac{d-c}{d \vee c}$ |

Each of the eight types of confirmation measures in Table 7 and Table 8 has its hypothesis-symmetrical or seccedent-symmetrical form. Therefore, there are 16 types of function $f(a, b, c, d)$ altogether for confirmation.

In prediction and converse prediction confirmation formulas, the conditions of conditional probabilities are the antecedents of rules so that the confirmation measures only depends on the two numbers of positive examples and counterexamples. Therefore, these measures accord with NFC.

If we change "$\vee$" into "+" in $f(a, b, c, d)$, measure $b^*$ becomes measure $b_F^* = F$, and measure $c^*$ becomes measure $c_F^*$. For example,

$$c_F^*(e_1 \rightarrow h_1) = (a\text{-}c)/(a+c). \tag{27}$$

# 4. Results

## 4.1. Relationship between measure $b^*$ and measure $F$

Measure $b^*$ is like measure $F$. Measure $b^*$ has all the above-mentioned desirable properties as well as measure $F$. The two measures changes with $LR$, as shown in Figure 6.

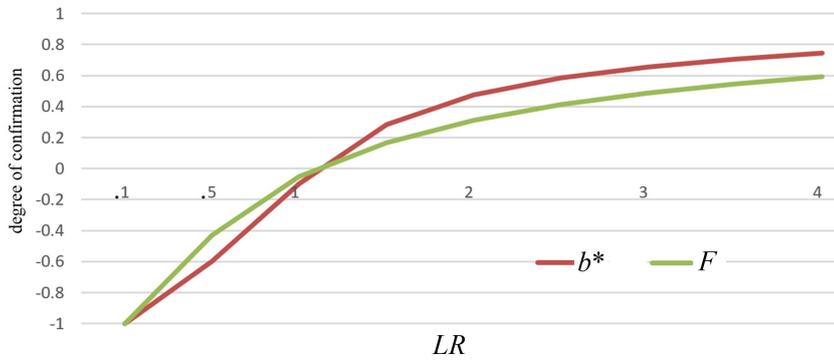

Fig. 6. Measures $b^*$ and $F$ change with $LR$

Measure $F$ has four confirmation formulas for different antecedents and seccedents [8]:

$$F(e_1 \rightarrow h_1) = \frac{P(e_1 \mid h_1) - P(e_1 \mid h_0)}{P(e_1 \mid h_1) + P(e_1 \mid h_0)} = \frac{ad - bc}{ad + bc + 2ac} = b_F^*(e_1 \rightarrow h_1), \tag{28}$$

$$F(h_1 \rightarrow e_1) = \frac{P(h_1 \mid e_1) - P(h_1 \mid e_0)}{P(h_1 \mid e_1) + P(h_1 \mid e_0)} = \frac{ad - bc}{ad + bc + 2ab} = b_F^*(h_1 \rightarrow e_1), \tag{29}$$

$$F(e_0 \rightarrow h_0) = \frac{P(e_0 \mid h_0) - P(e_0 \mid h_1)}{P(e_0 \mid h_0) + P(e_0 \mid h_1)} = \frac{ad - bc}{ad + bc + 2bd} = b_F^*(e_0 \rightarrow h_0), \tag{30}$$

$$F(h_0 \rightarrow e_0) = \frac{P(h_0 \mid e_0) - P(h_0 \mid e_1)}{P(h_0 \mid e_0) + P(h_0 \mid e_1)} = \frac{ad - bc}{ad + bc + 2cd} = b_F^*(h_0 \rightarrow e_0). \tag{31}$$

$F$ is equivalent to $b_F^*$ and similar to $b^*$.

## 4.2. Relationships between Prediction Confirmation Measures and some medical test's indexes

Channel confirmation measures are related to likelihood ratios, whereas Prediction Confirmation Measures (PCMs), including converse PCMs, are related to correct rates and false rates in the medical test.

To help us understand the significances of PCMs in the medical test, Table 9 shows that each PCM is related to which correct rate and which false rate.

Table 9. PCMs are related to correct rates and false rates in the medical test [27]

| PCM | Correct rate positively related to $c^*$ | False rate negatively related to $c^*$ |
|---|---|---|
| $c^*(e_1 \rightarrow h_1)$ | $P(h_1|e_1)$: positive predictive value | $P(h_0|e_1)$: false discovery rate |
| $c^*(e_0 \rightarrow h_0)$ | $P(h_0|e_0)$: negative predictive value | $P(h_1|e_0)$: false omission rate |
| $c^*(h_1 \rightarrow e_1)$ | $P(e_1|h_1)$: sensitivity or true positive rate | $P(e_0|h_1)$: false negative rate |
| $c^*(h_0 \rightarrow e_0)$ | $P(e_0|h_0)$: specificity or true negative rate | $P(e_1|h_0)$: false positive rate |

The false rates related to PCMs are the false report rates of the rule $e \rightarrow h$, whereas the false rates related to converse PCMs are the missing report rates of the rule $h \rightarrow e$.

### 4.3. Using three examples to compare various confirmation measures

We use two examples to explain that measure $b^*$ is more sensitive to counterexamples than to positive examples (see Figure 7).

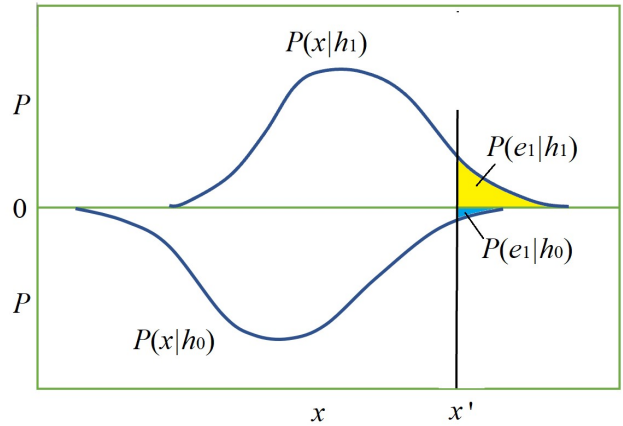

(a) Example 1: Positive examples' proportion is 0.1; Counterexample's proportion is 0.01.

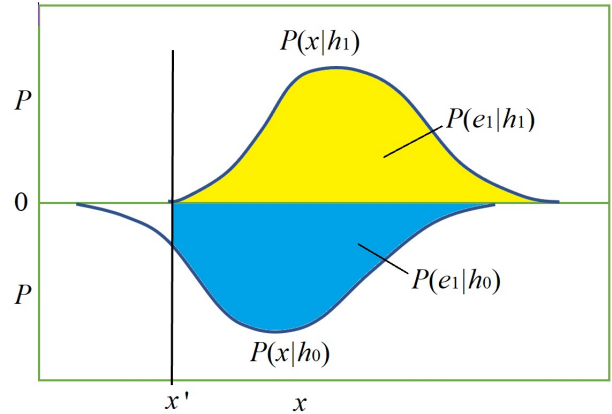

(b) Example 2: Positive examples' proportion is 1; Counterexample's proportion is 0.9.

Figure 7. How positive examples and counterexamples' proportions affect $b^*(e_1 \rightarrow h_1)$

In example 1, $b^*(e_1 \rightarrow h_1)=(0.1-0.01)/0.1=0.9$, which is very big. In example 2, $b^*(e_1 \rightarrow h_1)=(1-0.9)/1=0.1$, which is very small. The two examples indicate that fewer counterexamples' existence is more essential to $b^*$ than more positive examples' existence. $F$, $c^*$, and $c_F^*$ also possess this characteristic. However, most confirmation measures do not possess this characteristic.

We further suppose $P(h_1)=0.2$ and $n=1000$. The degrees of confirmation with different confirmation measures for the above two examples are shown in Table 10, where the base of log for $R$ and $L$ is 2. Table 10 also includes Example 3 (e.g., Ex. 3), whose $P(h_1)$ is 0.01. Example 3 reveals the difference between $Z$ and $b^*$ (or $F$).

Table 10. Three examples to show the differences between different confirmation measures.

| Ex. | a, b, c, d | D | M | R | C | Z | S | N | L | F | b* | c* |
|-----|------------|---|---|---|---|---|---|---|---|---|----|----|
| 1 | 20, 180, 8, 792 | .514 | <u>.072</u> | 1.84 | <u>.014</u> | .643 | .529 | <u>.09</u> | **3.32** | **.818** | **.9** | **0.8** |
| 2 | 200, 0, 720, 80 | .017 | <u>.08</u> | 0.12 | <u>.016</u> | .022 | .217 | <u>.1</u> | **.152** | **.053** | **.1** | **-.722** |
| 3 | 10, 0, 90, 900 | .09 | .9 | 3.32 | .009 | **.091** | .1 | .091 | 3.46 | **.833** | **.91** | **-.9** |

Data for Example 1 and 2 show that $L$, $F$ and $b*$ give Example 1 much higher rating than Example 2, whereas $M$, $C$, and $N$ give Example 2 higher rating than Example 1.

For Examples 2 and 3, in which $c>a$, only the values of $c*(e_1 \rightarrow h_1)$ are negative. The negative values should be reasonable for assessing probability predictions when counterexamples are more than positive examples.

The data for Example 3 show that when $P(h_0)=0.99>>P(h_1)=0.01$, $Z$ is very different from $F$ and $b*$ because $F$ and $b*$ are independent of $P(h)$.

### 4.4 How D, LR⁺, F, and c* are affected by Δa and Δd

Table 11 shows the degrees of confirmation calculated with four different measures. First, we suppose $a=d=20$ and $b=c=10$ to calculate the four degrees of confirmation. Next, we only replace $a$ with $a+\Delta a$ ($\Delta a =10$) to calculate the four degrees. Last, we only replace $d$ with $d+\Delta d$ ($\Delta d=10$) to calculate them.

Table 11. How $D$, $LR^+$, $F$, and $c*$ are affected by $\Delta a$ and $\Delta d$

| | f(a, b, c, d) | a=d=20 b=c=10 | Δ a=10 | Δ d=10 | Δf/Δd>Δf/Δa? |
|--|---------------|---------------|--------|--------|--------------|
| $D(e_1 \rightarrow h_1)$ | $a/(a+c)-(a+b)/n$ | 1/6 | 5/28 | 5/21 | Yes |
| $LR^+$ | $[a/(a+b)]/[c/(c+d)]$ | 2 | 9/4 | 3 | Yes |
| $F(e_1 \rightarrow h_1)$ | $(ad-bc)/(ad+bc+2ac)$ | 1/3 | 5/13 | 5/11 | Yes |
| $c*(e_1 \rightarrow h_1)$ | $(a-c)/\max(a, c)$ | 1/2 | 2/3 | 1/2 | No |

The cause of $\Delta f/\Delta d>\Delta f/\Delta a$ for $D$, $LR$, and $F$ is that $\Delta d$ can evidently decrease $P(h_1)$ and $P(e_1|h_0)$.

## 5. Discussions

### 5.1. To clarify the Raven Paradox (RP)

To clarify the RP, some researchers including Hemple [3] affirm the EC and deny the NFC; some researchers, such as Scheffler and Goodman [28], affirm the NFC and deny the EC. There are also some researchers do not fully affirm the EC or the NFC.

First we consider measure $F$ to see if we can use it to eliminate the RP. The difference between $F(e_1 \rightarrow h_1)$ and $F(h_0 \rightarrow e_0)$ is that their counterexamples are the same, yet, their positive examples are different (in Eqs. (28) and (31), $2ac$ and $2cd$ in the two denominators are different). As $d$ increases to $d+\Delta d$, $F(e_1 \rightarrow h_1)$ and $F(h_0 \rightarrow e_0)$ unequally increase. Therefore,

- though measure $F$ denies the EC, it still affirms that $F(e_1 \rightarrow h_1)$ and $F(h_0 \rightarrow e_0)$ are positively related;

- Measure $F$ does not accord the NFC.

Measure $b*$ is similar to $F$. The conclusion is that measures $F$ and $b*$ cannot eliminate our confusion about the RP.

After inspecting many different confirmation measures from the perspective of the rough set theory, Greco *et al.* [15] conclude that NC（e.g., NFC）is right, but it is difficult to find a suitable measure that accords with NC. However, many researchers still believe that the NC is incorrect; the NC accords with our intuition because $c(e_1 \rightarrow h_1)$ evidently increases with $a$ and slightly increases with $d$. After comparing different confirmation measures, Fitelson and Hawthorne [29] believe that the *LR* may be used to explain that a black raven can confirm more than a non-black non-raven thing.

Unfortunately, Table 11 shows that the increment of $D(e_1 \to h_1)$, $LR^+$, or $F(e_1 \to h_1)$ caused by $\Delta d$ is bigger than that caused by $\Delta a$ in this case. Therefore, these measures cannot be used to explain that a black raven can confirm much more than a non-black non-raven thing.

Nevertheless, measure $c*$ is different. Since $c*(e_1 \to h_1)=(a-c)/(a \lor c)$ and $c*(h_0 \to e_0)=(d-c)/(d \lor c)$, the EC does not hold, and measure $c*$ accords with the NFC very well. Hence, the RP does not exist anymore according to measure $c*$.

### 5.2. To distinguish statistical probability and logical probability

A hypothesis or a label's Logical Probability (LP) is the probability in which the hypothesis or the label is judged to be true, whereas its Statistical Probability (SP) is the probability in which it is selected. Suppose that ten thousand people go through a door. For everyone denoted by *x*, entrance guards judge "if *x* is elderly". If two thousand people are judged to be elderly, then the LP of the predicate "*x* is elderly" is 2000/10000=0.2. If the task of entrance guards is to select a label for everyone with one of four labels: "Child", "Youth", "Adult", and "Elderly", there may be one thousand people who are labeled "Elderly". The SP of "Elderly" should be 1000/10000=0.1. Why are not two thousand people are labeled "Elderly"? The reason is that some elderly people are labeled "Adult". LP of a tautology is 1, whereas its SP is almost 0 in general because a tautology is rarely selected. SP is normalized (the sum is 1), whereas LP is not normalized in general. Because of this reason, we should use two different symbols, such as "*P*" and "*T*", to distinguish SP and LP. The hybrid of LP and SP may produce subjective probability, such as $P(h|\theta_j)$ in Eq.(4). To calculate subjective or semantic information, we need subjective probability or LP [17]. To calculate degrees of confirmation, we need SP.

### 5.3. Selecting hypotheses and confirming rules: two tasks from the view of statistical learning

Researchers have noted the similarity between most confirmation measures and information measures. One explanation [30] is that information is the average of confirmatory impact. However, this paper gives a different explanation as follows.

There are three tasks in statistical learning: label learning, classification, and reliability analysis. There are similar tasks in inductive reasoning:

- 1) **Induction**. It is similar to label learning. For uncertain hypotheses, label learning is to train a likelihood function $P(x|\theta_j)$ or a truth function $T(\theta_j|x)$ by a sampling distribution [17]. The Logistic function often used for binary classifications may be treated as the truth function.

- 2) **Hypothesis selection**. It is similar to classification according to different criteria.

- 3) **Confirmation**. It is similar to reliability analysis. The classical methods are to provide likelihood ratios and correct rates (including false rates).

Classification and reliability analysis are two different tasks. Similarly, hypothesis selection and confirmation are two different tasks.

In statistical learning, classification depends on the criterion. The often-used criteria are the maximum posterior probability criterion, which is equivalent to the maximum correctness criterion, and the maximum likelihood criterion, which is equivalent to the maximum semantic information criterion [17]. The classifier for binary classifications is

$$e(x) = \begin{cases} e_1, & \text{if } P(\theta_1 \mid x) \ge P(\theta_0 \mid x), \text{or } P(x \mid \theta_1) \ge P(x \mid \theta_0); \\ e_0, & \text{otherwise.} \end{cases} \tag{32}$$

After the above classification, we may use information criterion to assess how well $e_j$ is used to predict $h_j$:

$$I(h_j;\theta_{ej})=I(h_j;e_j)=\log\frac{P(h_j\mid e_j)}{P(h_j)}=\log\frac{P(e_j\mid h_j)}{P(e_j)}$$

$$=\log P(h_j\mid e_j)-\log P(h_j)=\log P(e_j\mid h_j)-\log P(e_j) \tag{33}$$

$$=\log P(h_j,e_j)-\log[P(h_j)P(e_j)].$$

With information amounts $I(h_i;\theta_{ej})$ $(i, j=0,1)$, we can optimize the classifier:

$$e_j*=f(x)=\arg\max_{e_j}[P(h_0\mid x)I(h_0;\theta_{ej})+P(h_1\mid x)I(h_1;\theta_{ej})]. \tag{34}$$

The new classifier will result in new Shannon's channel $P(e|h)$. The maximum mutual information classification can be achieved by repeating Eqs. (33) and (34) [24,17].

With the above classifiers, we can make prediction $e_j$="$x$ is $h_j$" according to $x$. To tell information receivers how reliable the rule $e_j{\rightarrow}h_j$ is, we need the likelihood ratio ($LR$) to indicate how good the channel is or the correct rate to indicate how good the probability prediction is. Confirmation is similar. We need to provide a confirmation measure similar to $LR$, such as $F$ or $b*$, and a confirmation measure similar to the correct rate, such as $c*$ or $c_F*$. The difference is that the confirmation measures should change between -1 and 1.

The main distinction between predictions' information assessment and rules' confirmation is that information assessment uses an information measure to assess every prediction of using $e$ to predict $h$ separately (many times), whereas confirmation uses a confirmation measure and many examples with $e$ and $h$ as facts to assess a rule one time. To calculate subjective or semantic information, we need subjective probability or logical probability [17]. To calculate degrees of confirmation, we need statistical probability.

In statistical learning, information receivers make probability predictions according to $e_j$, $LR_j$, and $P(h)$ or according to $e_j$ and the corresponding correct rate. In inductive reasoning, the receivers make probability predictions according to $e_j$, $b*(e_j{\rightarrow}h_j)$, and $P(h)$, or according to $e_j$ and $c*(e_j{\rightarrow}h_j)$.

According to above analyses, it is easy to find that confirmation measures $D, N, R,$ and $C$ are more like information measures for assessing and selecting predictions instead of confirming rules. $Z$ is their normalization [8]; it seems between an information measure and a confirmation measure. However, confirming rules is different from assessing predictions; it needs the proportions of positive examples and counterexamples and is similar to calculating likelihood ratios or correct rates.

Confirmation is often explained as assessing the impact of the evidence on the hypothesis, or the impact of the premise on the consequence of a rule [14, 19]. However, this paper has a little different point of view that confirmation is to assess how well a sample or sampling distribution supports a major premise or a rule; the impact on the rule may be made by newly added examples (see the following section).

### 5.4. About incremental confirmation

Since one can use one or several examples to calculate the degree of confirmation with a confirmation measure, many researchers call their confirmation as incremental confirmation [14,15]. There are also researchers who claim that we need absolute confirmation [31]. This paper supports absolute confirmation.

The problem with incremental confirmation is that the degrees of confirmation calculated are often bigger than 0.5 and are irrelevant to our prior knowledge or $a, b, c, d$ that we knew before. It is unreasonable to ignore prior knowledge. Suppose that the logical probability of $h_1$="$x$ is elderly" is 0.2; the evidence is $x$=65 or x>60; the conditionally logical probability of $h_1$ is 0.9. With measure $D$, the degree of confirmation is 0.9-0.2=0.7, which is very big.

In function $f(a, b, c, d)$ for $F$, $b*$ and $c*$, the numbers $a, b, c,$ and $d$ should be those of all examples including past and current examples. A measure $f(a, b, c, d)$ should be an absolute confirmation measure. Its increment should be

$$\triangle f = f(a+\triangle a,\ b+\triangle b,\ c+\triangle c,\ d+\triangle d) - f(a, b, c, d). \tag{35}$$

If only $d$ increases to $d+\Delta d$ and $d>>\Delta d$, then $\Delta f/\Delta d$ must be very small. For example, if there 10 black ravens, 1 non-black raven, and 100 non-black non-raven things, then we have $c^*(e_1 \rightarrow h_1)=(a-c)/(a\lor c)=0.9$ and $c^*(h_0 \rightarrow e_0)=(d-c)/(d\lor c)=0.99$. The increment $\Delta f$ caused by $\Delta a=1$ is $10/11-9/10=0.0091$; the increment $\Delta f$ caused by $\Delta d=1$ is $100/101-99/100=0.0001$. The former is 91 times the latter. It can be seen that the increment of the degree of confirmation brought about by a new example is closely related to the number of old examples.

The absolute confirmation requires that sample size $n$ is big enough, that each example is generated independently, and that all examples are representative. If $a$, $b$, $c$, and $d$ are very small, the degree of confirmation calculated is unreliable. We need to replace the degree of confirmation with the degree interval of confirmation, such as [0.5, 1] instead of 1.

### 5.5. To distinguish a major premise's evidence and its consequence's evidence

The evidence for the consequence of a syllogism is the minor premise, whereas the evidence for confirming a major premise or a rule is a sample or a sampling distribution $P(e, h)$. In some researchers' studies, $e$ is used sometimes as the minor premise, and sometimes as an example or a sample; $h$ is used sometimes as the consequence, and sometimes as the major premise. Researchers use $c(e, h)$ or $c(h, e)$ instead of $c(e \rightarrow h)$ because they need to avoid the contradiction between the two understandings. However, if we distinguish two types of evidence, it has no problem to use $c(e \rightarrow h)$. We only need to emphasize that the evidence for the major premise is the sampling distribution $P(e, h)$ instead of $e$.

If $h$ is used as a major premise and $e$ is used as the evidence (such as in [5]), $\neg e$ is puzzling because there are four types of examples instead of two. Suppose $h=p \rightarrow q$ and $e$ is one of $(p, q)$, $(p, \neg q)$, $(\neg p, q)$, and $(\neg p, q)$. If $(p, \neg q)$ is the counterexample and other three examples $(p, q)$, $(\neg p, q)$ and $(\neg p, \neg q)$ are positive examples, which support $p \rightarrow q$, then $(\neg p, q)$ and $(\neg p, \neg q)$ should also support $p \rightarrow \neg q$ because of the same reason. However, according to HS or SS, it is unreasonable that the same evidence supports both $p \rightarrow q$ and $p \rightarrow \neg q$. In addition, $e$ is a sample with many examples in general. A sample's negation or a sample's probability is also puzzling.

Fortunately, though many researchers say that $e$ is the evidence of a major premise $h$, they also treat $e$ and $h$ as the antecedent and the seccedent of a major premise because, only in this way, one can calculate the probabilities or conditional probabilities of $e$ and $h$ for a confirmation measure. Why, then, should we replace $c(e, h)$ with $c(e \rightarrow h)$ to make the task clearer? In the following, we can see that $h$ used as a major premise will result in misunderstanding the above symmetries.

### 5.6. Is Hypothesis Symmetry (**HS**) or Seccedent Symmetry (**SS**) desirable?

Elles and Fitelson defined HS by $c(e, h)=-c(e, \neg h)$. Actually, it means $c(x, y)=-c(x, \neg y)$ for any $x$ and $y$. Similarly, Evidential Symmetry (ES) is Antecedent Symmetry (AS), which means $c(x, y)=-c(\neg x, y)$ for any $x$ and $y$. Since $e$ and $h$ are not the antecedent and seccedent of a major premise from their point of view, they could not say AS and SS. Consider that $c(e, h)$ becomes $c(h, e)$. According to the literal meaning of HS, one may misunderstand HS as shown in Table 12.

Table 12. Misunderstood HS and ES

|  | HS or SS | ES or AS |
| --- | --- | --- |
| Misunderstood HS | $c(e, h)=-c(e, \neg h)$ | $c(h, e)=-c(\neg h, e)$ |
| Misunderstood ES | $c(h, e)=-c(h, \neg e)$ | $c(e, h)=-c(\neg e, h)$ |

For example, the misunderstanding happens in [8,19], where $c(h, e)=-c(h, \neg e)$ is called ES. However, it is actually HS [14] or SS. In [19], the authors think that $F(E, H)$ (where the right side is the evidence) should have HS, whereas $F(E, H)$ should have ES. However, this "ES" does not accord with the original meaning of ES in [14]. Actually, both $F(H, E)$ and $F(E, H)$ possess HS or SS. The more serious thing because of the misunderstanding is that [18] concludes that ES and EHS (e.g., $c(H, E)=c(\neg H, \neg E)$), as well as HS, are desirable, and hence measures $S$, $N$, and $C$ are particularly valuable.

The author of this paper approves the conclusion of Elles and Fitelson about symmetries and asymmetries that only HS (SS) is desirable. Therefore, it is necessary to make clear that $e$ and $h$ in $c(e, h)$ are the antecedent and seccedent of the rule $e \rightarrow h$. To avoid the misunderstanding, we had better replace $c(e, h)$ with $c(e \rightarrow h)$ and use "AS" and "SS" instead of "ES" and "HS" .

### 5.7. About Bayesian confirmation and Likelihoodist confirmation

Measure $D$ proposed by Carnap is often referred to as the standard Bayesian confirmation measure. The above analyses, however, show that $D$ is only suitable as a measure of selecting hypotheses instead of a measure of confirming major premises. Carnap opened the direction of Bayesian confirmation, but his explanation of $D$ easily lets us confuse a major premise's evidence (a sample) and a consequence's evidence (a minor premise).

Greco *et al.* [19] call confirmation measures with conditional probability $p(h|e)$ Bayesian confirmation measures, those with $P(e|h)$ Likelihoodist confirmation measures, and those for $h \rightarrow e$ converse Bayesisn/Likelihoodist confirmation measures. This division is very enlightening. However, the division of confirmation measures in this paper does not depend on symbols, but on methods. The optimized proportion of the believable part in the truth function is the channel confirmation measure $b^*$, which is similar to the Likelihood Ratio (LR), reflecting how good the channel is; the optimized proportion of the believable part in the likelihood function is the prediction confirmation measure $c^*$, which is similar to the correct rate, reflecting how good the probability prediction is. The $b^*$ may be called the logical Bayesian confirmation measure (because it is derived with Logical Bayesian Inference [17]), although $P(e|h)$ may be used for $b^*$. The $c^*$ may be regarded as the likelihoodist confirmation measure, although $P(h|e)$ may be used for $c^*$.

This paper also provides converse channel/prediction confirmation measures for rule $h \rightarrow e$. Confirmation measures $b^*(e \rightarrow h)$ and $c^*(e \rightarrow h)$ are related to false report rates, whereas converse confirmation measures $b^*(h \rightarrow e)$ and $c^*(h \rightarrow e)$ are related to missing report rates.

### 4.8. About the certainty factor for probabilistic expert systems

The Certainty Factor (CF) was proposed by Shortliffe and Buchanan for a backward chaining expert system MYCIN [7]. It indicates how true an uncertain inference $h->e$ is. The relationship between the CF and the measure Z is $CF(h \rightarrow e) = Z(e \rightarrow h)$ [32].

As pointed out by Heckerman and Shortliffe [32], the CF method has been widely adopted in rule-based expert systems, it also has its theoretical and practical limitations. The main reason is that the CF method is not compatible with statistical probability theory. They believe that the belief-network representation can overcome many of the limitations of the CF model, that the CF model is simpler, and that it is possible to combine both to develop simpler probabilistic expert systems.

Measure $b^*(e_1->h_1)$ is related to the believable part of the truth value of proposition $e_1(h_1)$. It is similar to $CR(h \rightarrow e)$. The differences are that $b^*(e_1->h_1)$ is independent of $P(h)$, where as $CR(h \rightarrow e)$ is related to $P(h)$, and that $b^*(e_1->h_1)$ is compatible with statistical probability theory, whereas $CR(h \rightarrow e)$ is not.

Is it possible to use measure $b^*$ or $c^*$ as the certainty factor to simplify belief-networks or probabilistic expert systems? This question is worth to be explored.

### 5.9. How confirmation measures F, b*, and c* are compatible with Popper's falsification thought

Popper affirms that a counterexample can falsify a universal hypothesis or a major premise. However, for an uncertain major premise, how do counterexamples affect its degree of confirmation? Confirmation measures $F$, $b^*$, and $c^*$ can reflect the importance of counterexamples. Table 10 contains two examples. In Example 1, the proportion of positive examples is small, and the proportion of counterexamples is still less so that the degree of confirmation is big. This example shows that to improve the degree of confirmation, it is not necessary to increase the conditional probability $P(e_1|h_1)$ (for $b^*$) or $P(h_1|e_1)$ (for $c^*$). In Example 2, although the proportion of positive examples is big, the proportion of counterexamples is not small, so that the degree of confirmation is very small. This example shows that to raise degree of confirmation, it is not sufficient to increase the posterior probability. It is necessary and sufficient to decrease the relative proportion of counterexamples.

Popper affirms that a counterexample can falsify a universal hypothesis, which can be explained that for the falsification of a universal hypothesis, it is essential to have no counterexample. Now for the confirmation of an uncertainly universal hypothesis, we can explain that it is essential to have fewer counterexamples. It can be seen that confirmation measures $F$, $b^*$, and $c^*$ are compatible with Popper's falsification thought.

Scheffler and Goodman [28] proposed selective confirmation based on Popper's falsification thought. They believe that black ravens support "raven→black" because they undermine "raven → non-black", and hence, support the opposite hypothesis "raven→black". Their reason why non-black ravens support "raven→non-black" is because non-black ravens undermine the opposite hypothesis "raven→black". However, they did not provide the corresponding confirmation measure. Measure $c^*(e_1→h_1)$ is what they need.

## 6. Conclusions

Using the semantic information and statistical learning methods and taking the medical test as an example, this paper has derived two confirmation measures $b^*(e→h)$ and $c^*(e→h)$. The measure $b^*$ is similar to the measure $F$ proposed by Kemeny and Oppenheim, and can reflect the channel characteristics of the medical test like Likelihood Ratio (LR), indicating how good a testing means is. Measure $c^*(e→h)$ is similar to the correct rate but varies between - 1 and 1. Both $b^*$ and $c^*$ can be used for probability predictions, and $b^*$ is suitable for predicting the probability of disease when the prior probability of disease is changed. Measures $b^*$ and $c^*$ possess symmetries and asymmetries proposed by Elles and Fitelson [14], monotonicity proposed by Greco *et al.* [15], normalizing property (between -1 and 1) suggested by many researchers.

This paper has shown that measure $c^*$ definitely denied the Equivalence Condition and exactly reflected Nicod-Fisher Criterion, and hence, could be used to eliminate the Raven Paradox. The new confirmation measures $b^*$ and $c^*$ as well as $F$ indicate that fewer counterexamples' existence was more essential than more positive examples' existence, therefore, $b^*$ and $c^*$ were compatible with Popper's falsification thought. The new confirmation measures support absolute confirmation instead of incremental confirmation.

It is possible to use measure $b^*$ or $c^*$ as the certainty factor to simplify belief-networks, which is worth to be explored. When the sample is small, the degree of confirmation calculated by any confirmation measure is not reliable, and hence, the degree of confirmation should be replaced with the degree interval of confirmation. The specific method needs further studies.

**Supplemental Material**: Excel File for Data in Table 10 and Table 11

This file can be downloaded from http://survivor99.com/lcg/Table10-11.xlsx. We can test different confirmation measures by changing *a, b, c*, and *d*.

## References

[1]  R. Carnap, Logical Foundations of Probability, second ed., University of Chicago Press, Chicago, 1962.
[2]  K. Popper, Conjectures and Refutations. Routledge, London and New York, 1963.
[3]  C. G. Hempel, Studies in the Logic of Confirmation, Mind 54 (213) (1945)1–26 and 97–121.
[4]  J. Nicod, Le problème logique de l'induction, Paris: Alcan. (Engl. transl. The logical problem of induction, in: *Foundations of Geometry and Induction*, London: Routledge, 2000.) , 1924, P. 219.
[5]  H. Mortimer, The Logic of Induction, Prentice Hall, Paramus, 1988.
[6]  P. Horwich, Probability and Evidence, Cambridge University Press, Cambridge, 1982.
[7]  E. H. Shortliffe, B. G. Buchanan, A model of inexact reasoning in medicine, Mathematical Biosciences 23 (3-4) (1975) 351–379.
[8]  V. Crupi, K. Tentori, M. Gonzalez, On Bayesian measures of evidential support: Theoretical and empirical issues, Philos. Sci. 74 (2) (2007) 229–252.
[9]  D. Christensen, Measuring confirmation, Journal of Philosophy 96 (9) (1999) 437– 461.
[10] R. Nozick, Philosophical Explanations, Clarendon, Oxford, 1981.
[11] I. J. Good, The best explicatum for weight of evidence, Journal of Statistical Computation and Simulation 19 (4) (1984) 294–299.
[12] J. Kemeny, P. Oppenheim, Degrees of factual support, Philosophy of Science 19 (4) (1952) 307–324.
[13]  B. Fitelson, Studies in Bayesian confirmation theory. Ph.D. Thesis, University of Wisconsin, Madison. 2001.
[14] E. Eells, B. Fitelson, Symmetries and asymmetries in evidential support. Philosophical Studies 107 (2) (2002) 129–142.
[15] S. Greco, Z. Pawlak, R. Slowiński, Can Bayesian confirmation measures be useful for rough set decision rules? Engineering Applications of Artificial Intelligence 17 (4) (2004) 345–361.

[16] S. Greco, R. Slowiński, I. Szczęch, Properties of rule interestingness measures and alternative approaches to normalization of measures, Information Sciences 216 (20) (2012) 1-16.

[17] C. Lu, Semantic information G theory and Logical Bayesian Inference for machine learning, Information, 10 (8) (2019) 261.

[18] I. R. Thornbury, D. G. Fryback, W. Edwards, Likelihood ratios as a measure of the diagnostic usefulness of excretory urogram information. Radiology 114 (3) (1975) 561–565.

[19] S. Greco, R. Slowiński, I. Szczęch, Measures of rule interestingness in various perspectives of confirmation, Information Sciences 346-347 (2016) 216–235.

[20] C. Lu, A generalization of Shannon's information theory, Int. J. of General Systems, 8 (6)(1999) 453-490.

[21] K. Tentori, V. Crupi, N. Bonini, Osherson D. Comparison of Confirmation Measures. Cognition 103 (1) (2007) 107-119.

[22] D. H. Glass, Entailment and symmetry in confirmation measures of interestingness, Information Sciences 279 (2014) 552–559.

[23] R. Susmaga, I. Szczęch, Selected group-theoretic aspects of confirmation measure symmetries, Information Sciences 346–347 (2016) 424-441.

[24] C. Lu, Semantic channel and Shannon channel mutually match and iterate for tests and estimations with maximum mutual information and maximum likelihood. in: *Proceedings of the 2018 IEEE International Conference on Big Data and Smart Computing*, Shanghai, China, 15 January 2018; IEEE Computer Society Press Room, Washington, 2018, pp. 15–18.

[25] C. E. Shannon, A mathematical theory of communication, Bell System Technical Journal 27 (1948) 379–429, 623–656.

[26] OraQuick, Available online: http://www.oraquick.com/Home (accessed on 31 December 2016).

[27] T. Fawcett, An Introduction to ROC Analysis. Pattern Recognition Letters. 27 (8) (2006) 861–874.

[28] I. Scheffler, N. J. Goodman, Selective confirmation and the ravens: a reply to Foster, Journal of Philosophy, 69 (3) (1972) 78-83.

[29] B. Fitelson, J. Hawthorne, How Bayesian confirmation theory handles the paradox of the ravens, in E. Eells, J. Fetzer (eds.), *The Place of Probability in Science*, Springer, Dordrecht, 2010, pp. 247–276.

[30] V. Crupi, K. Tentori, State of the field: Measuring information and confirmation, Studies in History and Philosophy of Science 47(C) (2014) 81-90.

[31] F. Huber, What Is the Point of Confirmation? Philosophy of Science 72(5) (2005)1146-1159.

[32] D. E. Heckerman, E. H. Shortliffe, From certainty factors to belief networks, Artificial Intelligence in Medicine, 4(1) (1992) 35-52.